\documentclass[letterpaper, 10 pt, journal, twoside]{IEEEtran}

\IEEEoverridecommandlockouts                              

\usepackage{amsmath} 
\usepackage{amssymb} 
\usepackage{amsfonts}
\usepackage{mathtools}
\usepackage{times}  
\usepackage{balance}
\usepackage{pdfpages}

\usepackage{tabularx}

\usepackage{graphicx} 
\usepackage{blindtext}
\usepackage{multicol}
\usepackage[bookmarks=true]{hyperref}
\usepackage{multirow}
\usepackage{tabularx}
\usepackage{booktabs}
\usepackage{subfig}
\usepackage{float}
\usepackage{color}
\usepackage{cite}
\usepackage[flushleft]{threeparttable}

\usepackage{xhfill}
\usepackage[ruled]{algorithm2e}

\definecolor{axesRed}{RGB}{255,0,0}
\definecolor{axesGreen}{RGB}{44,160,44}
\definecolor{axesBlue}{RGB}{0,0,255}
\definecolor{myGold}{RGB}{255,192,0}

\title{
On Bundle Adjustment for Multiview Point \\ Cloud Registration
}

\author{Huaiyang Huang$^1$, Yuxiang Sun$^2$, Jin Wu$^1$, Jianhao Jiao$^1$, Xiangcheng Hu$^1$, Linwei Zheng$^1$, \\Lujia Wang$^1$ and Ming Liu$^1$
\thanks{
    Manuscript received March 25, 2021; revised July 5, 2021; accepted June 29, 2021. This paper was recommended for publication by Editor Pauline Pounds upon evaluation of the Associate Editor and the Reviewers’ comments.
    This work was supported by Zhongshan Municipal Science and Technology Bureau Fund, under project ZSST21EG06, Collaborative Research Fund by Research Grants Council Hong Kong, under Project No. C4063-18G, and Department of Science and Technology of Guangdong Province Fund, under Project No. GDST20EG54, awarded to Prof. Ming Liu.
}
\thanks{$^{1}$The authors are with the Robotics Institute at the Hong Kong University of Science and Technology, Hong Kong.
}
\thanks{
$^{2}$Yuxiang Sun is with the Department of Mechanical Engineering at the Polytechnic University, Hong Kong.
}
}

\begin{document}



\newcommand\Tstrut{\rule{0pt}{2.6ex}}         
\newcommand\Bstrut{\rule[-0.9ex]{0pt}{0pt}}   
\newcommand\todo[1]{\textcolor{yellow}{TODO: #1.}}         

\IEEEaftertitletext{\vspace{-1.5\baselineskip}}

\markboth{IEEE ROBOTICS AND AUTOMATION LETTERS. PREPRINT}{HUANG \MakeLowercase{\textit{et al.}}: Bundle Adjustment}
\maketitle

\begin{abstract}
    Multiview registration is used to estimate Rigid Body Transformations (RBTs) from multiple frames and reconstruct a scene with corresponding scans.
    Despite the success of pairwise registration and pose synchronization, the concept of Bundle Adjustment (BA) has been proven to better maintain global consistency.
    So in this work, we make the multiview point-cloud registration more tractable from a different perspective in resolving range-based BA.
    Based on this analysis, we propose an objective function that takes both measurement noises and computational cost into account.
    For the feature parameter update, instead of calculating the global distribution parameters from the raw measurements, we aggregate the local distributions upon the pose update at each iteration.
    The computational cost of feature update is then only dependent on the number of scans.
    Finally, we develop a multiview registration system using voxel-based quantization that can be applied in real-world scenarios.
    The experimental results demonstrate our superiority over the baselines in terms of both accuracy and speed. Moreover, the results also show that our average positioning errors achieve the centimeter level. Related materials are available at our project page \texttt{\url{https://hyhuang1995.github.io/bareg/}}.
\end{abstract}
\vspace{-1.5em}


\IEEEpeerreviewmaketitle

\section{Introduction}
\label{sec:intro}


\IEEEPARstart{P}{oint} cloud registration has long been an essential research problem in the computer vision and robotics community \cite{liu2020hercules}.
Particularly for autonomous navigation, the availability of range sensors capable of efficiently reconstructing the 3D structure has significantly improved the systematic performance \cite{huang2020gmmloc}.
This fact stimulates the research on multiview registration, which combines multiple scans of point-cloud data (PCD) into a globally consistent structural model \cite{bergevin1996towards}.

In the early stage, the multiview registration problem is resolved by either performing pairwise registration \cite{faugeras1986representation, zhang2014loam}, or combining this scheme with pose synchronization \cite{rosen2019se}.
For decades, pairwise registration is the most prevailing paradigm and is widely applied in different state estimation systems for range sensors \cite{besl1992method,zhang2014loam}.
To tackle this problem in the multiview regime, the systems are generally designed in a frame-to-frame or frame-to-model fashion.
The problem is thus simplified into resolving the Rigid Body Transformation (RBT) concerning the latest scan.
Despite the prevalence and success of various pairwise registration methods,
global consistency is seldom considered in these implementations.

On the top of pairwise registration methods, some research work explicitly deals with the global consistency with the introduction of pose synchronization \cite{pulli1999multiview,huber2003fully}.
This is achieved by first registering pairs of overlapped scans and then optimizing a pose graph from the registered relative poses.
However, as the objective function is not directly optimized, the attainable level of consistency is restricted.
Despite significant progress, challenges remain for better range-based registration.
To this end, recent research resorts to the Bundle Adjustment (BA), a concept that originated from visual Structure-from-Motion (SfM) that simultaneously estimates the frame poses and the position of visual landmarks.
Nevertheless, compared to visual state estimation applications, relatively fewer discussions on leveraging inter-frame constraints can be found for systems based on range sensors.
Some methods are proposed to fully exploit the inter-frame constraints provided by a sequence of point-cloud data. 
Firstly, Landmark-based methods parameterize the local measurements as individual landmarks. Similar to visual BA, landmark and pose parameters are jointly optimized \cite{weingarten20063d, geneva2018lips}.
These methods generally require specific feature detection, and the measurement noises are not easy to be taken into account.
Most recently, two works propose to directly use the eigenvalue of sample covariance as the objective function \cite{ferrer2019eigen,liu2020balm}, which is denoted as EigenValue Minimization (EVM)-based formulation.
These methods provide an elegant formulation for multiview registration and build a system using the proposed method. However, the trade-off exists between the generality in formulation and performance in applications.


In this paper, we propose a formulation and a system for multiview registration.
To this end, we first analyze the optimal condition of EVM, which unifies the Landmark and the EVM-based method.
Secondly, based on the above analysis, we derive an objective function for resolving multiview registration, which can be applied to a standard least-square problem for more efficient optimization.
Different from previous methods, this formulation also fully considers the measurement noise.
Finally, we provide a systematic implementation that can be applied in real-world scenarios.
We summarize our contributions as follows:
\begin{enumerate}
	\item We provide an analysis on the optimal condition of EVM-formulation for resolving multiview registration, which unifies the methods by extracting landmarks and directly optimizing eigenvalues.
	\item We propose a novel objective function based on the analysis, which takes account of both computational efficiency and the measurement noise.
	\item We develop a voxel-based multiview registration pipeline with the proposed objective function and local distribution aggregation.
\end{enumerate}



\section{Related Works}
\label{sec.review}


\subsection{Pairwise Registration}
Pairwise registration is the most prevailing paradigm in the range-based state estimation.
This track of methods considers the only variable to optimize is the pose of the latest frame.
To estimate, for example, the states of a sequence of point-cloud data, they generally can be divided into two categories, i.e.,
frame-to-frame and frame-to-model methods.


Frame-to-frame registration has long been an important problem in robotic perception.
Among these works, ICP \cite{besl1992method} is one of the most popular pipelines, along with its variants \cite{rusinkiewicz2001efficient,pomerleau12comp}.
In an Expectation-Maximum (EM) scheme, they find the correspondences with the pose estimation and optimize the pose parameters with the association result.
As an extension to frame-to-frame method, the frame-to-model pipeline is widely applied in SLAM \cite{zhang2014loam,shan2018lego,ye2019tightly} and localization \cite{zhu2019real,rozenberszki2020lol} systems for robotic state estimation.
For example, in LOAM \cite{zhang2014loam}, scans in the past frames are aggregated into a global map with their estimated poses.
After every frame-to-frame registration step, the system refines the pose estimation with the downsampled global map.
Some work \cite{zhang2014loam, qin2020lins} shows that the frame-to-model refinement significantly improves the localization and mapping performance.
Despite the success and brevity of these methods, the concept of BA is seldom exploited, unlike vision-based pipelines. 
 





\subsection{Multiview Registration}

In contrast to pairwise registration, approaches targeting multiview registration take multiple frames of PCD that are partially overlapped as input. With local observations cross different frames, they simultaneously optimize the poses of different frames in multiple point-cloud frames.

In the early stage, the majority of methods are based on pairwise registration and pose synchronization \cite{pulli1999multiview, bergevin1996towards, lu1997globally,  mendes2016icp, huang2019learning}.
This formulation is not directly derived from the sensor measurement model, thus it is more suitable for producing initial estimation for further fine-tuning.
Another track of methods parameterizes the measurements as different geometric landmarks \cite{weingarten20063d, kaess2015simultaneous,geneva2018lips}, namely the Landmark-base methods.
Similar to previous methods, the formulation does not consider the measurement noise,
and they generally require extracting geometric features in advance.

Most recently, some work provides a different perspective to this problem \cite{ferrer2019eigen, liu2020balm}.
They directly optimize the eigenvalues of the local distribution parameters, yielding an elegant and global consistent solution.
In this paper, we further discuss the optimal condition of this formulation.
However, different from previous works, we provide a novel objective function that can be applied to a least-squares solver with better convergence speed.
\section{Background}
\label{sec.bg}

\subsection{Notations}

We use bold uppercase $\mathbf{H}$ for the matrices, bold lowercase $\mathbf{x}$ for the vector, and light lower case (e.g. $\theta$) for the scalar.
For a matrix $\mathbf{H}$, its eigenvalues are denoted by $ \lambda_i \left(\mathbf{H}\right) $, which are enumerated in a descending order: $\lambda_1 \geqslant \dots \geqslant \lambda_n$.
For $\mathbf{H} \in \mathbb{R}^{3 \times 3}$ and $\mathbf{H}$ is a symmetric matrix,
$\lambda_i$ can be calculated from Singular Value Decomposition (SVD) by $\mathbf{H} = \mathbf{R}_{\mathbf{H}}\boldsymbol{\Lambda}_{\mathbf{H}}{\mathbf{R}_{\mathbf{H}}}^T$, where we have
\begin{equation}
    \boldsymbol{\Lambda}_{\mathbf{H}} = \text{diag}(\lambda_1, \lambda_2, \lambda_3),
     \mathbf{R}_{\mathbf{H}} \in \text{SO}(3).
\end{equation}


The pose of $k$-th frame $\mathcal{F}_k$ is represented by the RBT $\mathbf{T}_k = \left(\mathbf{R}_k, \mathbf{t}_k\right)$ , where $\mathbf{R}_k \in \text{SO}(3)$ and $\mathbf{t}_k \in \mathbb{R}^3$.
A 3D point $\mathbf{p}_i$ expressed under $\mathcal{F}_k$ can be transformed to the global frame $\mathcal{W}$ using: $\mathbf{R}_k \mathbf{p}_i + \mathbf{t}_k$.
$\mathbf{T}_k$ can be further parameterized as a vector via $\boldsymbol{\xi}_k = \log\left({\mathbf{T}_k}^\lor\right), \boldsymbol{\xi}_k \in \mathfrak{se}(3)$ and expressed back via $\mathbf{T}_k = \exp\left({\boldsymbol{\xi}_k}^\land\right)$.

\subsection{Problem Formulation}



Suppose a common feature is observed by a set of frames $\left\{\mathcal{F}_1, \dots, \mathcal{F}_N\right\}$ and the feature is parameterized by $\boldsymbol{\pi}$ under $\mathcal{W}$.
Then the total state vector can be defined by $\mathbf{x} = \left[\boldsymbol{\xi}_1, \dots \boldsymbol{\xi}_N, \boldsymbol{\pi}\right]$.
The local observation of this feature in $\mathcal{F}_k$ is represented by a set of points $\mathcal{P}_k = \left\{\mathbf{p}_{k1}, \dots, \mathbf{p}_{kn_k} \right\}$, and the geometric constraint is modeled by the measurement function: $r\left(\boldsymbol{\xi}_k, \mathbf{p}_{ki}, \boldsymbol{\pi} \right)$.
Taking the planar feature as an example, if we parameterize the planar feature as $\boldsymbol{\pi} = \left[\mathbf{n}, \boldsymbol{\mu}\right]$,
where $\mathbf{n}$ is the normal of the plane and $\boldsymbol{\mu}$ can be an arbitrary point on the surface.
For the simplicity of the following discussions, we define $\boldsymbol{\mu}$ as the sample mean.
Accordingly, the geometric constraint is the point-to-plane distance:
\begin{equation}
    r\left(\boldsymbol{\xi}_k, \mathbf{p}_{ki}, \boldsymbol{\pi} \right) =
    \mathbf{n}^T \left(\mathbf{R}_k\mathbf{p}_{ki} + \mathbf{t}_k - \boldsymbol{\mu}\right).
\end{equation}
With the Gaussian noise assumption, the objective of multiview registration problem is to minimize the following energy function:
\begin{equation}
    \hat{\mathbf{x}} = \arg \min_{\mathbf{x}}  \sum_{k=1}^{N} \frac{1}{n_k}\sum_{i=1}^{n_k} \left\|
    r\left(\boldsymbol{\xi}_k, \mathbf{p}_{ki}, \boldsymbol{\pi} \right)
    \right\|_2^2.
    \label{eq.obj}
\end{equation}
Methods for solving this optimization problem are described in the following section.


\subsection{Resolving Multiview Registration}

Here we provide a brief introduction to the previous methods as for the background of further discussions.

\textbf{Planar Landmark (PL)}
This track of methods estimates feature parameters from the local measurements and then simultaneously optimizes the feature and pose parameters.
This scheme is very similar to the pipeline of visual BA.
There are diverse methods for feature parameterization and estimation.
For example, Principle Component Analysis (PCA) is a widely used technique for planar feature extraction.
Given the set of local measurements $\left\{\mathbf{p}_{k1}, \dots, \mathbf{p}_{kn_k} \right\}$ under $\mathcal{F}_k$, sample mean and covariance is given as:
\begin{equation}
    \left\{
    \begin{aligned}
         & \boldsymbol{\mu}_k^\ell = \frac{1}{n_k}\sum_{i=1}^{n_k} \mathbf{p}_{ki} \\
         & \boldsymbol{\Sigma}_k^\ell = \frac{1}{n_k}\sum_{i=1}^{n_k}
        \left(\mathbf{p}_{ki} - \boldsymbol{\mu}_k^\ell\right)
        \left(\mathbf{p}_{ki} - \boldsymbol{\mu}_k^\ell\right)^T
    \end{aligned}
    \right. ,
\end{equation}
And the surface normal is estimated from Singular Value Decomposition (SVD) of $\boldsymbol{\Sigma}_k^\ell$.
To leverage inter-frame constraints, generally, a distance function on $\boldsymbol{\pi}$ is defined, which is appended to the objective function \autoref{eq.obj} in the optimization.


\textbf{BALM}
Given the current estimated poses, the points under global coordinate can be aggregated as
$\mathcal{P} = \left\{ \mathbf{R}_{ki}\mathbf{p}_{ki} + \mathbf{t}_k | k = 1,\dots,N , i = 1, \dots n_k\right\}$.
The sample mean and covariance of $\mathcal{P}$ are denoted by $\boldsymbol{\mu}$ and $\boldsymbol{\Sigma}$, given by:
\begin{equation}
    \left\{
    \begin{aligned}
         & \boldsymbol{\mu} = \frac{1}{\sum_{k} n_k}\sum_{i=1}^{n_k} \mathbf{p}_{ki}^\prime \\
         & \boldsymbol{\Sigma} = \frac{1}{\sum_k n_k}\sum_{i=1}^{n_k}
        \left(\mathbf{p}_{ki}^\prime - \boldsymbol{\mu}\right)
        \left(\mathbf{p}_{ki}^\prime - \boldsymbol{\mu}\right)^T
    \end{aligned}
    \right. ,
\end{equation}
where $\mathbf{p}_{ki}^\prime = \mathbf{R}_k \mathbf{p}_{ki} + \mathbf{t}_k$.

\textit{Lemma. 1}
Assume known optimal feature parameters $\boldsymbol{\pi}$, the objective function is equivalent to minimize the minimal eigenvalue, that is:
\begin{equation}
    \lambda_3\left(\boldsymbol{\Sigma}\right) =
    \frac{1}{\sum_k n_k}\sum_{i=1}^{n_k} \left\|
    \mathbf{n}^T \left(\mathbf{R}_k\mathbf{p}_{ki} + \mathbf{t}_k - \boldsymbol{\mu}\right)
    \right\|_2^2.
    \label{eq.a}
\end{equation}

\textit{Proof.} We refer the readers to BALM \cite{liu2020balm} and \cite{hyhuang2021basup}.

Based on \textit{Lemma. 1}, BALM directly resolves the multiview registration problem via:
\begin{equation}
    \hat{\mathbf{x}} = \arg \min_{\mathbf{x}} \lambda_3 \left(\boldsymbol{\Sigma}\right).
\end{equation}
We name this formulation as EigenValue Minimization (EVM) formulation.
Assuming the optimal feature parameter $\boldsymbol{\pi}$ is calculated in advance of the optimization, this formulation is only dependent on the frame poses.

\textbf{Eigen-Factor (EF)}
EF uses Homogenous point representation $\tilde{\mathbf{p}} = \left[\mathbf{p}, 1\right]$, and the local feature is parameterized as $\boldsymbol{\eta} = \left[\mathbf{n}, -\mathbf{n}^T \boldsymbol{\mu}\right]$.
The objective function is re-derived into:
\begin{equation}
    \hat{\mathbf{x}} = \arg \min_\mathbf{x} \sum_{k=1}^{N}
    \boldsymbol{\eta}^T \mathbf{T}_k
    \underbrace{
        \tilde{\mathbf{P}}_k
        \tilde{\mathbf{P}}_k^T
    }_{\mathbf{S}_k}
    \mathbf{T}_k^T
    \boldsymbol{\eta},
    \label{eq.ef}
\end{equation}
where each column of $\tilde{\mathbf{P}}_k \in \mathbb{R}^{4 \times n_k}$ corresponds to stacked transformed homogeneous points.
Then with the first-order gradient descending method, EF optimizes this objective function to resolve the frame poses.
Note that as this formulation gives the same point-to-plane distance,
it is equivalent to the EVM as in \autoref{eq.a}.






\textit{Remark. 1}
The first issue is about the feature parameter estimation in PL-based methods.
Introducing plane parameters into the optimization variables lead to a large optimization structure.
This also causes information loss as the raw measurements are neglected.

\textit{Remark. 2}
The first issue is the handling of the optimal feature parameters.
We observe that the optimal feature parameters $\hat{\boldsymbol{\pi}}$ varies with the update of frame poses $\left\{\mathbf{T}_k\right\}$.
As a consequence, although the EVM formulation is independent of feature parameters, the update of feature parameters would require recomputation of $\boldsymbol{\mu}$ and $\boldsymbol{\Sigma}$.
Therefore, BALM assumes the optimal feature parameter is resolved before optimization, and EF uses Homogeneous representation to avoid point-wise update of $\boldsymbol{\mu}$ and $\boldsymbol{\Sigma}$.
However, as the formulation of EF is applicable only to planar features, it restricts the discussion on other geometric feature types.

\textit{Remark. 3}
Another trade-off exists between the formulation and the efficiency.
As the original objective function is re-formulated as the minimal eigenvalue,
it is no longer in a least-squares formulation.
Solving it with first-order gradient descent methods is difficult to converge efficiently.
On the contrary, BALM uses a second-order approximation which requires the sophisticated computation of large-scale Hessian matrices, the time complexity of which is dependent on the number of points in the input scan.




\section{Methodology}
\label{sec.method}

In this section, we provide theoretical analysis for the optimal condition of \autoref{eq.a}.
Then we introduce an objective function that takes measurement noise into account and does not cause computational overhead.
Finally, we apply this formulation to a voxel-based multiview registration algorithm as an application example.






\subsection{On the Optimal Condition of EVM}

EVM actually provides an elegant formulation that can be considered as the basis of multiview registration.
Both BALM and EF are derived from \autoref{eq.obj}.
Therefore the optimality is self-evident.
Here we provide theoretical analysis for an interesting finding that Plane Landmark and EVM can actually be unified under certain conditions.
As the $\lambda_3(\boldsymbol{\Sigma})$ represents the ``thickness'' of the aggregated point clouds, our intuition is that if the local features share the same parameters, the ``thickness'' should be minimized.
Following this intuition, the optimal condition is given as:
\begin{equation}
    \left\{
    \begin{aligned}
         & \mathbf{R}_k \cdot (\mathbf{R}_{\boldsymbol{\Sigma}_k} \mathbf{e}_{z})
        \parallel
        \hat{\mathbf{n}}, \quad \forall k,                                                                                   \\
         & \left( \boldsymbol{\mu}_k - \boldsymbol{\mu}_{k^\prime} \right) \perp \hat{\mathbf{n}}, \quad \forall k, k^\prime
    \end{aligned}
    \right. ,
    \label{eq.opt_cond}
\end{equation}
where $\mathbf{R}_{\boldsymbol{\Sigma}_k}$ is from the decomposition result: $\boldsymbol{\Sigma} = \mathbf{R}_{\boldsymbol{\Sigma}_k}  \boldsymbol{\Lambda}_{\boldsymbol{\Sigma}_k}\mathbf{R}_{\boldsymbol{\Sigma}_k}^T$ and $\mathbf{e}_z = \left[0, 0, 1\right]^T$.








\begin{algorithm}[t]
    \SetAlgoLined
    \KwIn{ $\{\mathcal{P}_k\}$, $\{\bar{\boldsymbol{\xi}}_k\}$, $r$}
    \KwResult{$\{\hat{\boldsymbol{\xi}}_k\}$, $\hat{\boldsymbol{\pi}}$}

    \CommentSty{// initialize map with resolution $r$}

    $\mathcal{M}$ = initializeVoxelMap($r$)

    \For(){$\mathcal{P}_k$ in $\{\mathcal{P}_k\}$}{

        castCloudToMap($\mathcal{P}_k$, $\mathcal{M}$)
    }

    $\mathcal{V}$ = countActiveVoxels($\mathcal{M}$)

    \For(){$\mathcal{V}_k$ in $\mathcal{V}$}{
        checkInliers($\mathcal{V}_k$)
    }

    \While{not converged}{
        $\hat{\boldsymbol{\pi}}$ = updateFeatureParam($\hat{\boldsymbol{\xi}}_k$)

        $\mathbf{J}$, $\delta\mathbf{r}$ = computeJacobianAndResidual($\mathcal{V}$, $\hat{\boldsymbol{\pi}}$)

        $\{\hat{\boldsymbol{\xi}}_k\}$ = updateState($\mathbf{J}$, $\delta\mathbf{r}$)
    }
    \caption{Voxel-based Multiview Registration}
    \label{alg.impl}
\end{algorithm}



Next, we prove that this is actually the optimal condition that minimizes the objective function \autoref{eq.obj}.
To this end, our goal is to prove:
For the minimization problem in \autoref{eq.a}, a solution $\{\hat{\mathbf{T}}_1, \dots, \hat{\mathbf{T}}_{n_k} \}$ is optimal iff it satisfies \autoref{eq.opt_cond}.
We begin with the proof of sufficiency.


\textit{Lemma. 2 (Weyl's inequality)}
Given $\mathbf{M} = \mathbf{N} + \mathbf{R}$,
where $\mathbf{N}$ and $\mathbf{R}$ are $n\times n$ symmetric matrices,
the following inequality holds for $ 1 \leqslant i \leqslant n$:
\begin{equation*}
    \lambda_i \left(\mathbf{N}\right) + \lambda_n \left(\mathbf{R}\right)
    \leqslant \lambda_i \left(\mathbf{M}\right) \leqslant
    \lambda_i \left(\mathbf{N}\right) + \lambda_1 \left(\mathbf{R}\right).
\end{equation*}


In \autoref{sec.local_agg}, we show that $ \boldsymbol{\Sigma} = \sum_{k} \frac{n_k}{n} \left( \boldsymbol{\Sigma}_k + \boldsymbol{\Sigma}_{\boldsymbol{\mu}_k} \right)$, where $\boldsymbol{\Sigma}_k$ and $\boldsymbol{\Sigma}_{\boldsymbol{\mu}_k}$ are both symmetric semi-positive matrices and the detailed derivation can be found in \autoref{sec.local_agg}.
Then by applying Weyl's inequality on $\boldsymbol{\Sigma}$, we have
\begin{equation}
    \begin{aligned}
        \lambda_3 \left(\boldsymbol{\Sigma}\right)
         & \geqslant \sum_{k} \frac{n_k}{n} \left(\lambda_3 \left(\boldsymbol{\Sigma}_k\right)  +  \lambda_3 \left(\boldsymbol{\Sigma}_{\boldsymbol{\mu}_k}\right) \right) \\
         & \geqslant \sum_{k} \frac{n_k}{n} \lambda_3 \left(\boldsymbol{\Sigma}_k\right).
    \end{aligned}
\end{equation}
This inequality gives a lower bound of $\lambda_3\left(\boldsymbol{\Sigma}\right)$.
Next we show that if \autoref{eq.opt_cond} is satisfied, the energy function reaches the lower bound.

\textit{Theorem. 1} If the optimal conditions \autoref{eq.opt_cond} are satisfied, the following equality holds:
\begin{equation}
    \boldsymbol{\Sigma}\hat{\mathbf{n}}
    =  \left(\sum_{k} \frac{n_k}{n} \lambda_3\left(\boldsymbol{\Sigma}_k\right)\right) \hat{\mathbf{n}}.
\end{equation}

\textit{Proof.}
By multiplying $\hat{\mathbf{n}}$ to both sides, we have
\begin{equation}
    \begin{aligned}
        \boldsymbol{\Sigma}\hat{\mathbf{n}} &
        = \sum_{k} \frac{n_k}{n}
        \left(\boldsymbol{\Sigma}_k \hat{\mathbf{n}} + \boldsymbol{\Sigma}_{\boldsymbol{\mu}_k} \hat{\mathbf{n}}\right) = \sum_{k} \frac{n_k}{n}
        \left(\boldsymbol{\Sigma}_k \hat{\mathbf{n}} \right)                                                                                      \\
                                            & = \sum_{k} \frac{n_k}{n}
        \left(\boldsymbol{\Sigma}_k \mathbf{R}_k \cdot (\mathbf{R}_{\boldsymbol{\Sigma}_k} \mathbf{e}_{z}) \right) \quad (\text{collinerity})     \\
                                            & = \sum_{k} \frac{n_k}{n}
        \left( \mathbf{R}_k\mathbf{R}_{\boldsymbol{\Sigma}_k}\boldsymbol{\Lambda}_k
        \underbrace{\mathbf{R}_{\boldsymbol{\Sigma}_k}^T\mathbf{R}_{k}^T \mathbf{R}_k \mathbf{R}_{\boldsymbol{\Sigma}_k}}_{\mathbf{I}_{3\times3}}
        \mathbf{e}_{z} \right)                                                                                                                    \\
                                            & = \sum_{k} \frac{n_k}{n} \lambda_3\left(\boldsymbol{\Sigma}_k\right)
        \mathbf{R}_k \mathbf{R}_{\boldsymbol{\Sigma}_k} \mathbf{e}_{z}                                                                            \\
                                            & =  \left(\sum_{k} \frac{n_k}{n} \lambda_3\left(\boldsymbol{\Sigma}_k\right)\right) \hat{\mathbf{n}}. \\
    \end{aligned}
\end{equation}
Based on Theorem 1, $\sum_{k} \frac{n_k}{n} \lambda_3\left(\boldsymbol{\Sigma}_k\right)$ is the eigenvalue of $\boldsymbol{\Sigma}$ and its corresponding eigenvector is $\hat{\mathbf{n}}$.
In addition, with the conditions satisfied, the lower bound of $\lambda_3 \left(\boldsymbol{\Sigma}\right)$ is reached.
This shows that \autoref{eq.opt_cond} is sufficient for the optimality.
Then, we raise a counter example to prove the necessity.

\textit{Definition. 1 (Rayleigh quotient)}
Given a symmetric matrix $\mathbf{A}$, the Rayleigh quotient is defined by:
\begin{equation}
    R(\mathbf{A}, \mathbf{x}) = \frac{\mathbf{x}^T \mathbf{A} \mathbf{x}}{\mathbf{x}^T \mathbf{x}}.
\end{equation}

\textit{Lemma. 3 (Bounds of Rayleigh quotient)} 
For a $3\times 3$ symmetric matrix $\mathbf{A}$ and vector $\mathbf{x}$, the bounds of Rayleigh quotient are given by:
\begin{equation}
    \lambda_3 \left(\mathbf{A}\right)
    \leqslant
    R(\mathbf{A}, \mathbf{x})
    \leqslant
    \lambda_1 \left(\mathbf{A}\right)
\end{equation}

Suppose we have a solution that does not satisfy \autoref{eq.opt_cond},
we show that we can always achieve a lower value of Rayleigh quotient by perturbing the motion parameters.
Given an optimal solution $ \hat{\mathbf{x}}$, and there exists $\mathbf{R}_k \mathbf{R}_{{\Sigma}_k} \mathbf{e}_z\not\perp$.
We perturb $\mathbf{R}_k$ by $\mathbf{R}_k^\prime = \mathbf{R}_k \delta \mathbf{R}_k$, so that we have $\mathbf{R}_k^\prime \mathbf{R}_{\boldsymbol{\Sigma}_k} \mathbf{e}_z \perp \hat{\mathbf{n}}$.
Accordingly, we show that the updated Rayleigh quotient is given as:
\begin{equation*}
    \begin{aligned}
        \hat{\mathbf{n}}^T\boldsymbol{\Sigma} \hat{\mathbf{n}}
         & = \frac{n_p}{n} \hat{\mathbf{n}}^T\boldsymbol{\Sigma}_k \hat{\mathbf{n}}+ c
        > \frac{n_p}{n} \hat{\mathbf{n}}^T\boldsymbol{\Sigma}_p \left(\mathbf{R}_p^\prime\right) \hat{\mathbf{n}}+ c,
    \end{aligned}
\end{equation*}
which does not reach the lower bound. Recall \textit{Lemma. 3}, $\hat{\mathbf{n}}$ is not the optimal solution, thus the necessity is proved.

\textit{Remark. 4}
The above proof implicitly unifies the theory of PL-based and EVM-based methods.
However, two factors position PL-based methods against other solutions.
The first is that the above proof assumes $\mathbf{n}$ is optimal under current estimation.
For PL-based methods, as the plane parameter is optimized based on local parameterization, there is no guarantee that it is the optimal solution.
The second issue is that certain information loss is encountered, especially with the measurement noise.
To handle the measurement noise properly, we re-formulate the equation with a weighting scheme described in the next section.

\subsection{Implementation on Multiview Registration}

\subsubsection{Objective Function}

Starting from the above proof, we now provide the proposed objective function based on the idea of both PL-based and EVM-based methods, which is given by:
\begin{equation}
    \begin{aligned}
        \hat{\mathbf{x}} = \arg \min_{\mathbf{x}} & \sum_{k=1}^N
        n_k \lambda_1 (\boldsymbol{\Sigma}_{k}) \| \mathbf{R}_k\mathbf{R}_{\boldsymbol{\Sigma}_k} \mathbf{e}_x \cdot \mathbf{n} \|^2                                                            \\
                                                  & + \sum_{k=1}^N n_k \lambda_2 (\boldsymbol{\Sigma}_{k}) \| \mathbf{R}_k\mathbf{R}_{\boldsymbol{\Sigma}_k} \mathbf{e}_y \cdot \mathbf{n} \|^2 \\
                                                  & + \sum_{k=1}^N n_k \| \mathbf{n}^T (\mathbf{R}_k \boldsymbol{\mu}_k+ \mathbf{t}_k - \boldsymbol{\mu}) \|^2,
    \end{aligned}
\end{equation}
where $\mathbf{R}_{\boldsymbol{\Sigma}_k}$ is decomposed in advance of the optimization.
In the formulation, the first two terms constrain the rotational component of individual poses, and the last term constrains both.
This gives a more clear interpretation geometrically.
In the implementation, $\mathbf{n}$ is supposed to be optimal under the current estimation, and this is actually not included in the optimization update.
After each step, we re-calculate $\mathbf{n}$ with aggregated covariance $\boldsymbol{\Sigma}$.
This formulation differs from the previous methods that 1) compared to Landmark-based methods, all the measurements are taken into account along with the noise (e.g., $\lambda_i$); 2) the residuals are naturally formed in a least-squares fashion, allowing efficient Hessian matrix approximation;
3) without loss of generality compared to the Homogeneous representation;
4) the objective function is only related to frame poses, thus there is no need for point-wise computation or downsampling the input point-cloud.

\begin{figure*}[t!]
    \centering
    \includegraphics[width=0.32\textwidth]{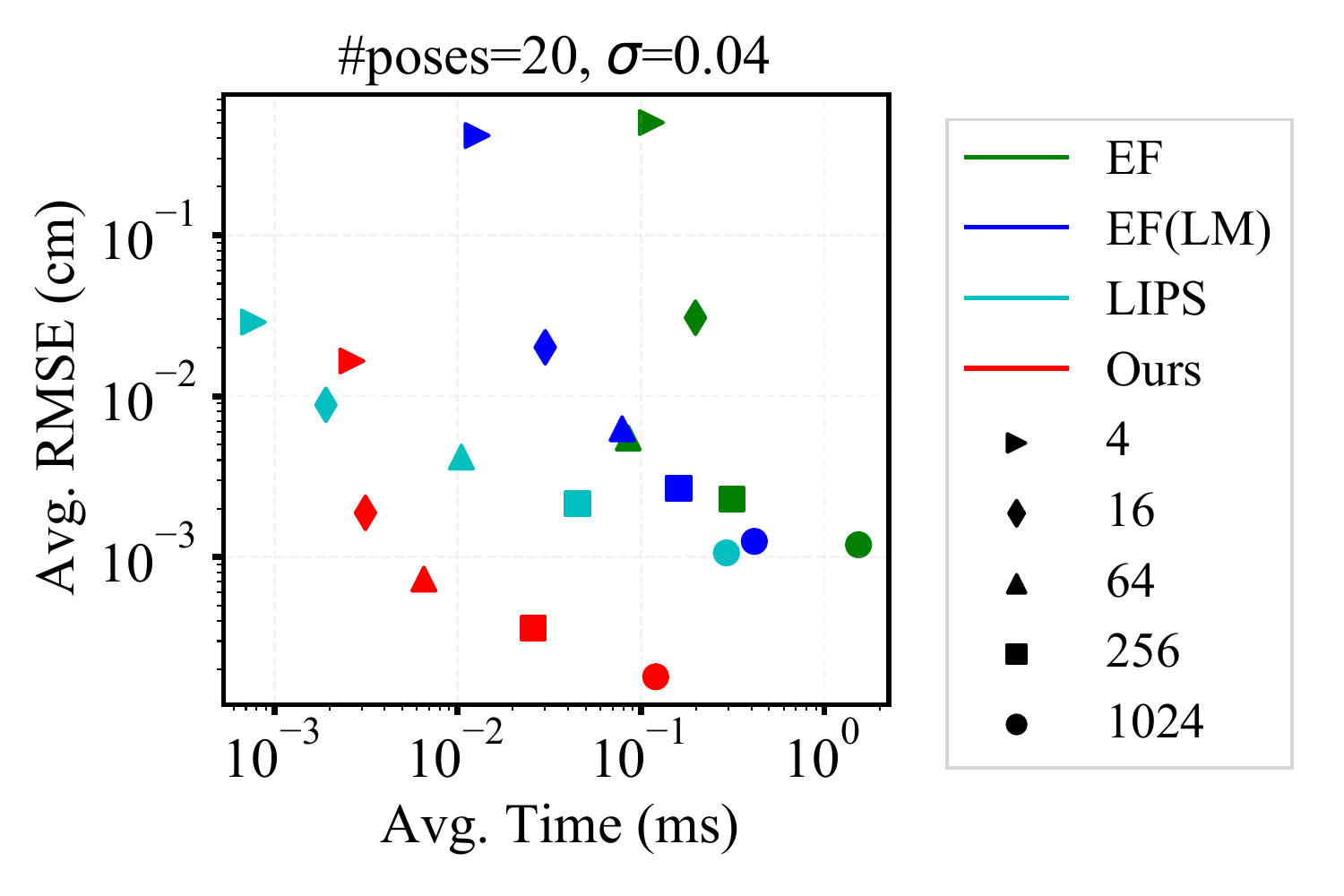} \,
    \includegraphics[width=0.32\textwidth]{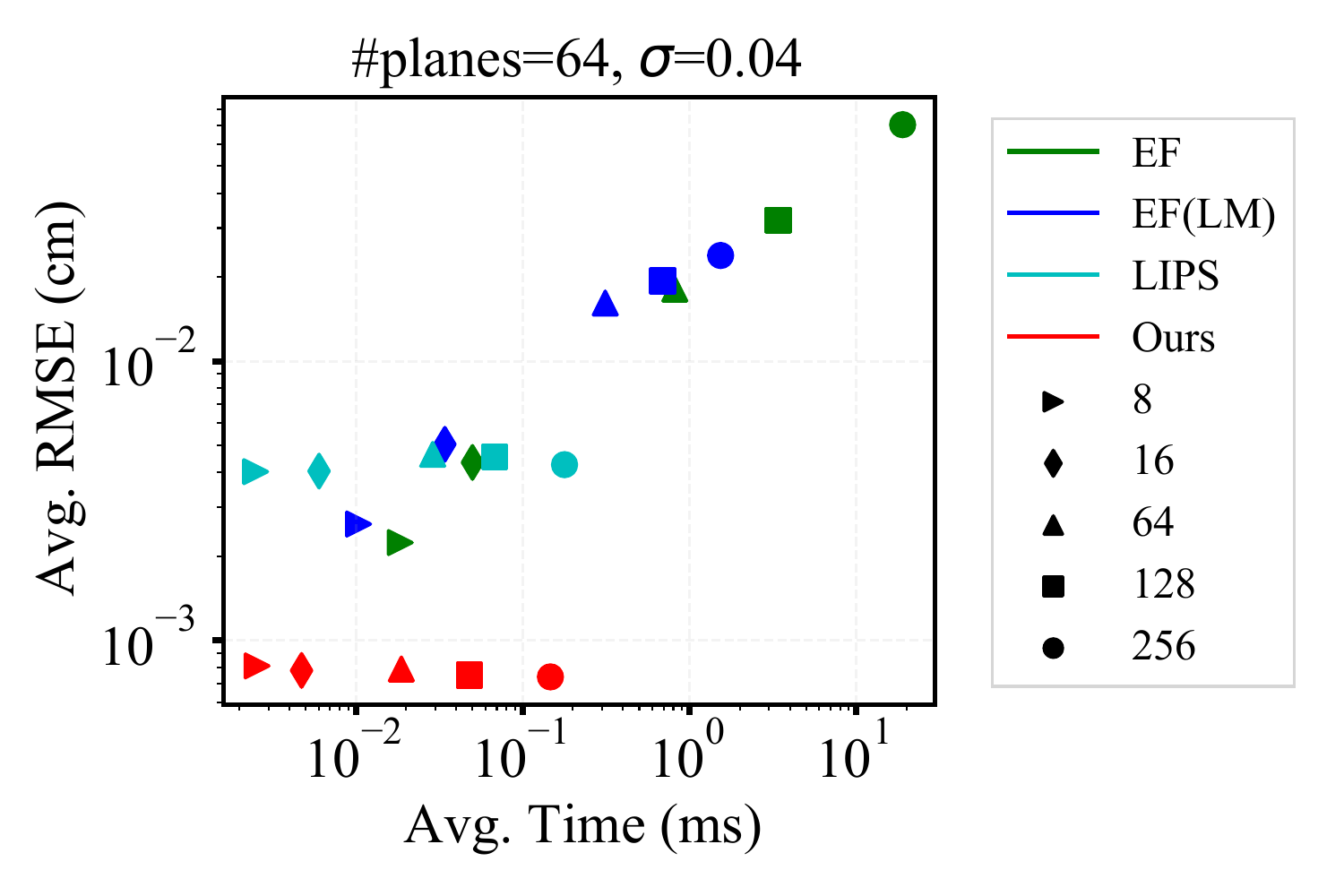}\,
    \includegraphics[width=0.32\textwidth]{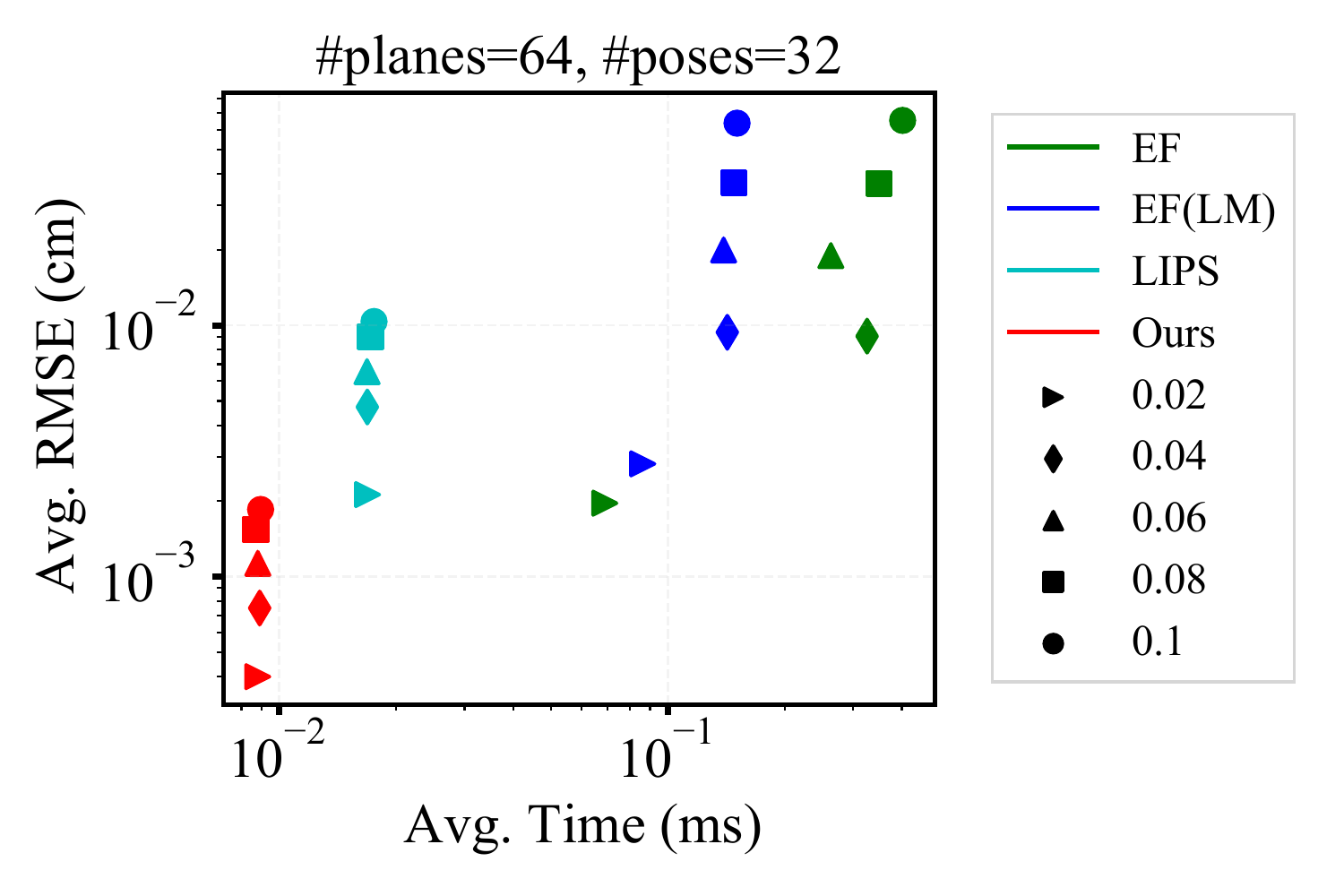} \\
    \vspace{0.5em}
    \includegraphics[width=0.32\textwidth]{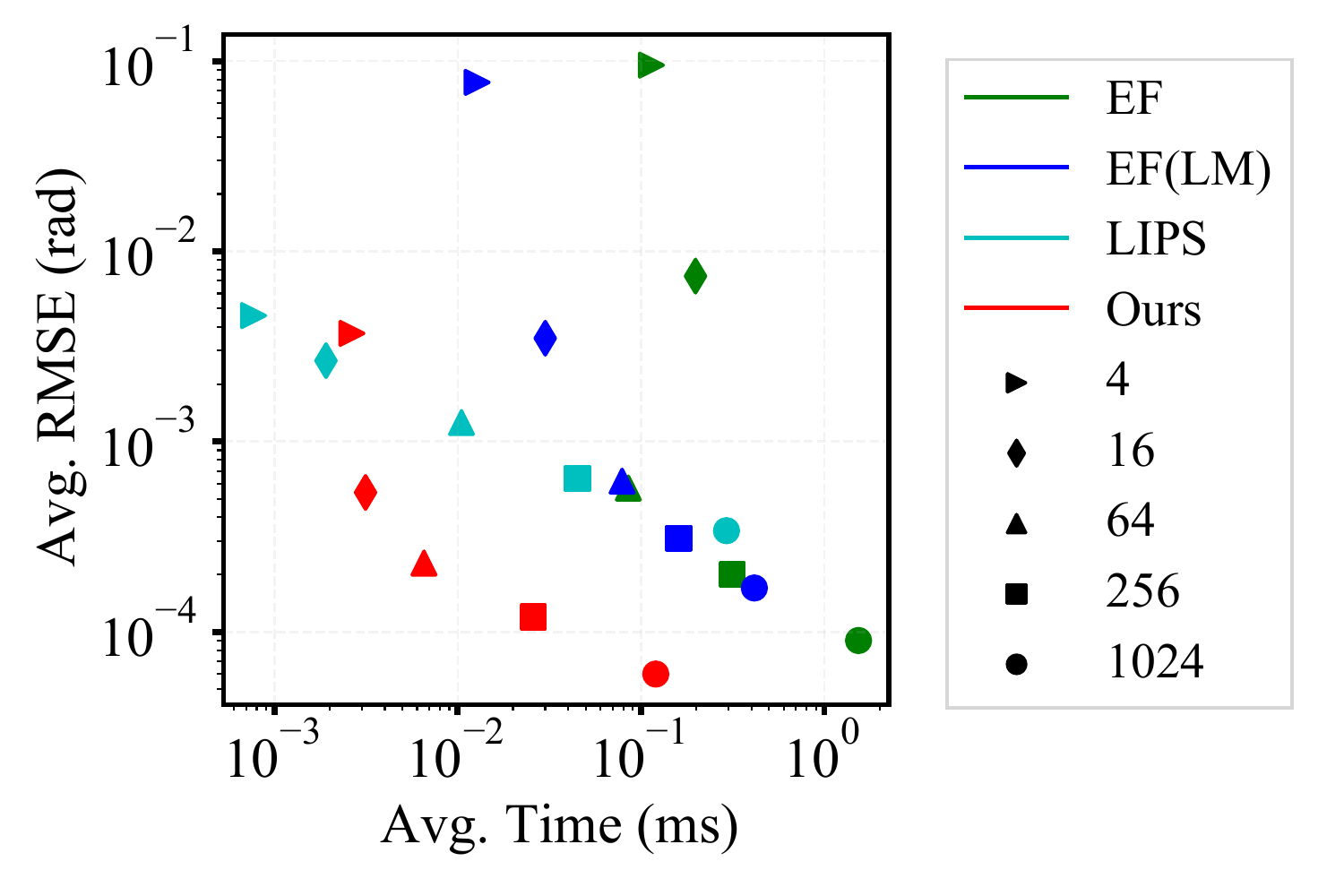} \,
    \includegraphics[width=0.32\textwidth]{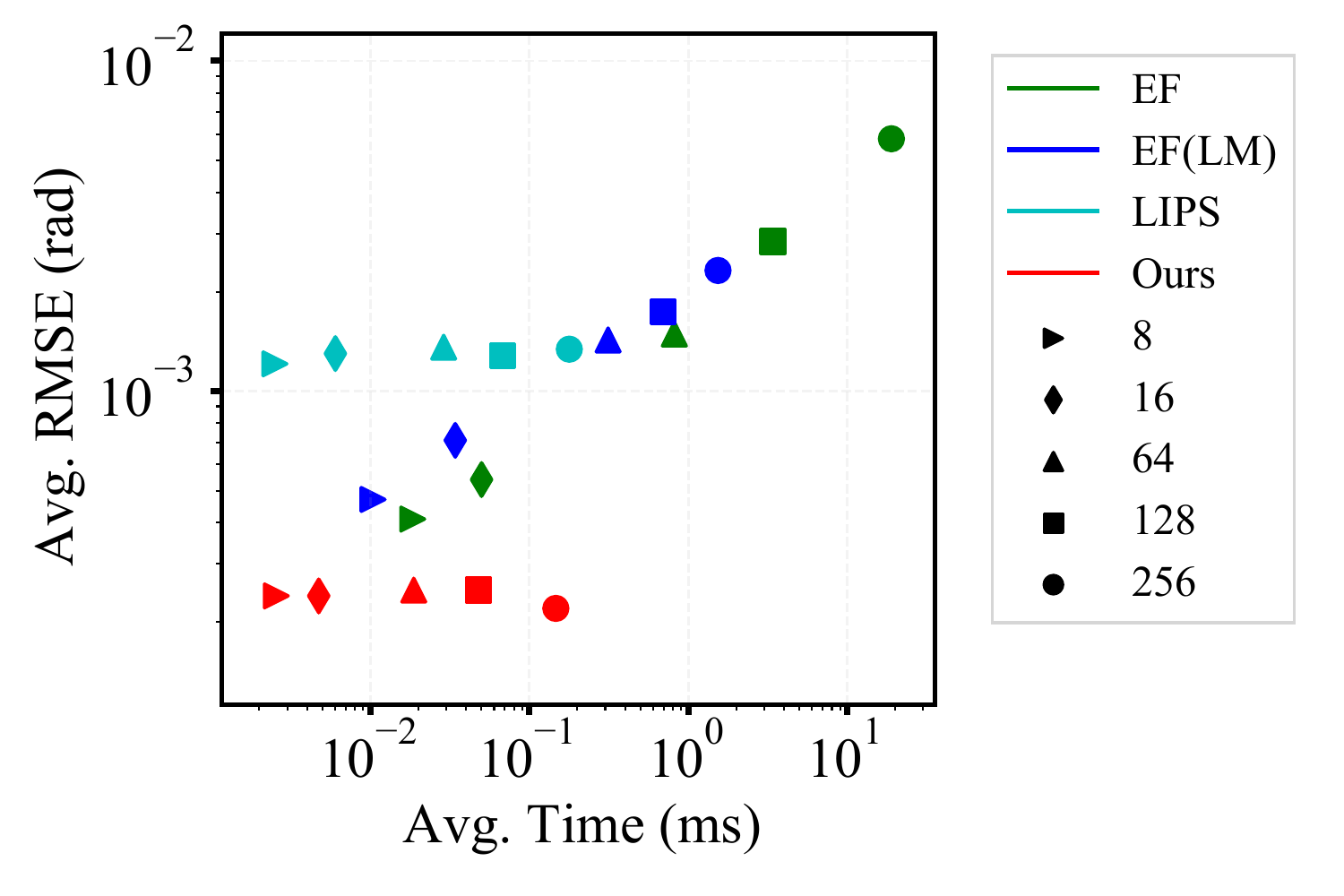}\,
    \includegraphics[width=0.32\textwidth]{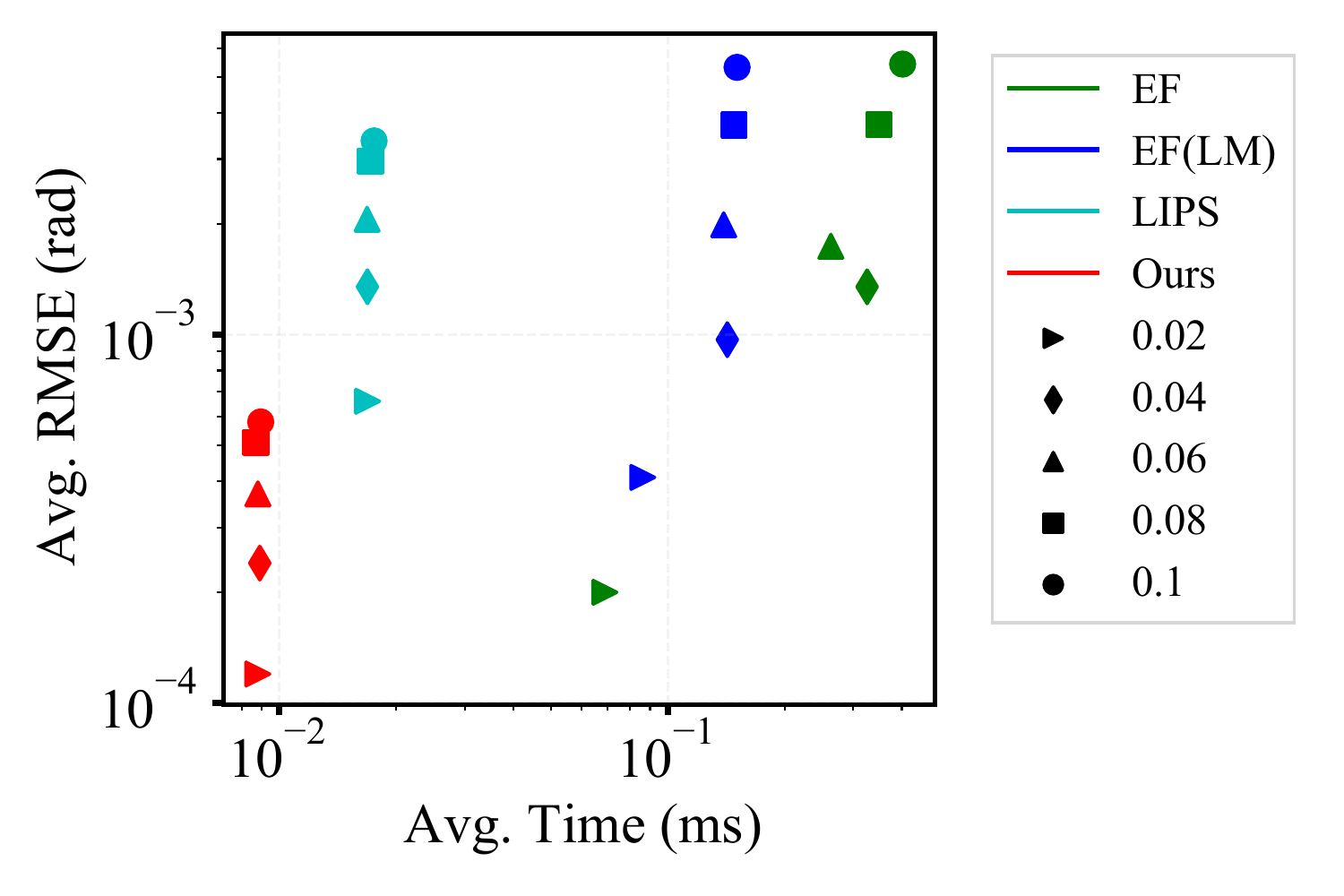}
    \caption{
        Speed-accuracy tests in simulation for different BA methods, where the y-axis is the corresponding translational (Top Row) and rotational (Bottom Row) error, with respect to different number of landmarks (Left Column), number of poses (Middle Column), and noise level (Right Column).
    }
    \label{fig.sim_err}
\end{figure*}

\subsubsection{Local Distribution Aggregation}
\label{sec.local_agg}


Another major concern of previous work is mentioned in \textit{Remark. 1}, i.e., the computational cost of feature parameter updating.
When pose update is performed, the update of sample mean and covariance require significant computation.
Here we derive a close-formed solution for the $\boldsymbol{\mu}$ and $\boldsymbol{\Sigma}$.
Moreover, the time complexity is then \textit{independent} of the total number of points $\sum_k n_k$ and only dependent on the number of frames $N$.

We can first estimate $\boldsymbol{\mu}_k^\ell$ and $\boldsymbol{\Sigma}_k^\ell$ under $\mathcal{F}_k$.
Given the corresponding pose of $\mathcal{F}_k$, the sample mean and covariance under $\mathcal{W}$ can be derived by linear transformation:
\begin{equation}
    \left\{
    \begin{aligned}
         & \boldsymbol{\mu}_k = \mathbf{R}_k \boldsymbol{\mu}_k^\ell + \mathbf{t}_k       \\
         & \boldsymbol{\Sigma}_k = \mathbf{R}_k \boldsymbol{\Sigma}_k^\ell \mathbf{R}_k^T
    \end{aligned}
    \right. ,
\end{equation}
With the estimation results of different subsets, the distribution for aggregated point cloud can be derived in closed form, given by:
\begin{equation}
    \left\{
    \begin{aligned}
         & \boldsymbol{\mu} = \sum_{k} \frac{n_k}{n} \boldsymbol{\mu}_k \\
         & \boldsymbol{\Sigma} = \sum_{k} \frac{n_k}{n}
        \left(
        \boldsymbol{\Sigma}_k
        + \boldsymbol{\Sigma}_{\boldsymbol{\mu}_k}
        \right)
    \end{aligned}
    \right. ,
\end{equation}
where the second term $\boldsymbol{\Sigma}_{\boldsymbol{\mu}_k}$ is given by:
\begin{equation}
    \boldsymbol{\Sigma}_{\boldsymbol{\mu}_k} =
    (\boldsymbol{\mu}_k -\boldsymbol{\mu})
    (\boldsymbol{\mu}_k -\boldsymbol{\mu})^T.
\end{equation}
For the simplicity, the detailed derivation is provided in \cite{hyhuang2021basup}.
With this formulation, the update of feature parameters can be efficient,
as the point-wise update is avoided.

\subsubsection{A Voxel-based Association}


To tackle a specific multiview registration problem, we implement a voxel-based multiview registration pipeline, which additionally deals with association and feature selection.
The detailed implementation of this system is illustrated in the algorithm. 1.
In the beginning, we build up a voxel map where the voxels are stored in a hash table with unique indices.
The voxel map is established with a specific resolution $r$.
Given each frame's point-cloud data, we then cast each frame into the voxel map with the initial frame pose.
We assume that each voxel corresponds to a global feature, and scanned points inside this voxel of a specific frame are considered the local observation of this feature.
The correspondences across different frames are associated according to this voxel representation.
Before the optimization, we determine active voxels and inlier local observations by:
\begin{enumerate}
    \item the number of local measurements is sufficient;
    \item the ratio between eigenvalues is appropriate.
\end{enumerate}
We aggregate the local distributions in each voxel with the updated frame poses at each optimization step, and then we estimate the feature parameters with the aggregated covariance $\boldsymbol{\Sigma}$.
After that, we first compute Jacobians and residuals, and apply the LM method to calculate parameter update $\delta \mathbf{x}$.
We iterate until the optimization converges, and finally, the optimal poses for different frames are calculated.

\section{Experimental Results and Discussions}
\label{sec.exp}


To evaluate the feasibility and effectiveness of the proposed method, we perform:
(1) evaluation on different BA methods on simulation.
(2) registration experiment on real-world scenarios.

\subsection{Monte Carlo Simulation}
To validate the proposed method against other formulations, we perform extensive Monte Carlo simulations on different methods for multiview registration.
\begin{table*}[t!]
    \caption{Evaluation of registration performance on ETHZ Registration Dataset. We report averaged Relative Pose Error (RPE [cm]) and Absolute Pose Error (APE [cm]) for different methods. The {\color{axesRed}{\textbf{best}}} and {\color{axesBlue}{\textit{second best}}} results are highlighted.}
    \begin{center}
        \footnotesize
        \begin{tabular}{l@{\hspace{1ex}}l@{\hspace{1ex}}| c@{\hspace{1ex}}|c@{\hspace{1ex}}|c@{\hspace{1ex}}|c@{\hspace{1ex}}|c@{\hspace{1ex}}|c@{\hspace{1ex}}|c@{\hspace{1ex}}|c@{\hspace{1ex}}|c@{\hspace{1ex}}}
            \toprule
            \multicolumn{2}{c}{\textbf{Method}}                            & Apart.                                                         & Haupt.                                                              & Stairs                                                         & Moun.                                                            & Gaze.S.                                                          & Gaze.W.                                                           & Wood.S                                                            & Wood.A                                                         & \textbf{Average}                                                            \\
            \midrule
            Global                                                         & TEASER                                                         & 5.5 / 15.1                                                          & 2.7 / 10.2                                                     & 4.3 / 10.0                                                       & 6.9 / 36.6                                                       & 3.4 / 8.0                                                         & 2.8 / 5.5                                                         & 4.2 / 7.3                                                      & 3.2 / 8.8                            & 4.1 / 12.7                           \\
                                                                           & FGR                                                            & 4.9 / 22.0                                                          & 8.7 / 86.4                                                     & 4.6 / 6.2                                                        & 8.6 / 30.3                                                       & 3.8 / 7.3                                                         & 4.5 / 12.7                                                        & 3.6 / 10.8                                                     & 3.3 / 7.0                            & 5.2 / 22.8                           \\ \midrule
            Global+Sync                                                    & TEASER                                                         & 4.1 / 10.7                                                          & 2.0 / 3.1                                                      & 3.7 / 5.6                                                        & 6.8 / 13.7                                                       & 3.2 / 3.4                                                         & 2.8 / 2.3                                                         & 5.1 / 4.6                                                      & 3.6 / 4.7                            & 3.9 / 6.0                            \\
                                                                           & FGR                                                            & 5.4 / 17.5                                                          & - / -                                                          & 4.3 / 8.1                                                        & 10.4 / 28.7                                                      & 4.1 / 6.2                                                         & 6.0 / 9.8                                                         & 5.5 / 19.0                                                     & 4.3 / 21.4                           & 6.5 / 27.2                           \\ \midrule
            F2F
                                                                           & ICP(pt2pt)                                                     & 2.1 / 9.4                                                           & 1.5 / 7.0                                                      & 1.5 / 3.9                                                        & 3.4 / 26.4                                                       & 1.4 / 6.6                                                         & 1.1 / 5.5                                                         & 2.3 / 10.0                                                     & 1.9 / 14.9                           & 1.9 / 10.5                           \\
                                                                           & ICP(pt2pl)                                                     & 1.2 / 3.9                                                           & {\color{axesRed}\textbf{0.4}} / 1.5                            & {\color{axesBlue}\textit{0.8}} / 1.8                             & {\color{axesRed}\textbf{2.1}} / 14.8                             & 1.1 / 4.4                                                         & {\color{axesBlue}\textit{0.9}} / 5.1                              & 2.5 / 9.3                                                      & 1.8 / 13.4                           & {\color{axesBlue}\textit{1.3}} / 6.8 \\
                                                                           & GICP                                                           & - / -                                                               & {\color{axesBlue}\textit{0.5}} / 1.2                           & {\color{axesRed}\textbf{0.7}} / 1.7                              & 4.8 / 45.1                                                       & 4.8 / 57.2                                                        & 2.3 / 16.9                                                        & 2.6 / 9.7                                                      & 2.0 / 13.0                           & 3.1 / 47.8                           \\
                                                                           & VGICP                                                          & {\color{axesRed}\textbf{0.6}} / {\color{axesRed}\textbf{1.1}}       & {\color{axesRed}\textbf{0.4}} / {\color{axesRed}\textbf{0.6}}  & 0.9 / 1.6                                                        & 35.5 / 84.3                                                      & 11.0 / 37.7                                                       & - / -                                                             & 3.0 / 7.7                                                      & 1.8 / 8.5                            & 10.9 / 49.2                          \\
                                                                           & NDT                                                            & 3.1 / 11.5                                                          & 2.5 / 6.2                                                      & 3.3 / 8.0                                                        & 4.3 / 21.3                                                       & 3.2 / 6.3                                                         & 3.1 / 9.0                                                         & 3.8 / 13.2                                                     & 3.4 / 20.1                           & 3.3 / 12.0                           \\ \midrule
            F2M
                                                                           & ICP(pt2pt)                                                     & 1.8 / 4.5                                                           & 1.1 / 2.0                                                      & 1.5 / 3.5                                                        & 3.8 / 23.2                                                       & {\color{axesBlue}\textit{1.0}} / {\color{axesBlue}{\textit{1.8}}} & {\color{axesRed}\textbf{0.7}} / {\color{axesBlue}\textit{1.3}}    & {\color{axesRed}\textbf{1.8}} / {\color{axesBlue}\textit{2.6}} & {\color{axesRed}\textbf{1.1}} / 4.9  & 1.6 / 5.5                            \\
                                                                           & ICP(pt2pl)                                                     & 3.1 / 30.2                                                          & {\color{axesBlue}\textit{0.5}} / {\color{axesRed}\textbf{0.6}} & {\color{axesRed}\textbf{2.1}} / 4.4                              & {\color{axesRed}\textbf{2.1}} / {\color{axesBlue}\textit{8.6}}   & 1.7 / 2.2                                                         & {\color{axesRed}\textbf{0.7}} / 1.4                               & 2.4 / 3.2                                                      & 1.5 / 4.7                            & 1.8 / 6.9                            \\
                                                                           & GICP                                                           & 3.9 / 35.5                                                          & 0.6 / {\color{axesRed}\textbf{0.6}}                            & {\color{axesBlue}\textit{0.8}} / {\color{axesRed}{\textbf{1.3}}} & 4.2 / 21.6                                                       & 1.1 / 2.7                                                         & {\color{axesRed}\textbf{0.7}} / {\color{axesBlue}\textit{1.3}}    & 2.1 / 2.5                                                      & 1.5 / {\color{axesBlue}\textit{4.3}} & 1.8 / 8.7                            \\
                                                                           & VGICP                                                          & - / -                                                               & 1.2 / 2.4                                                      & 16.1 / 24.5                                                      & - / -                                                            & 2.1 / 2.0                                                         & 1.8 / 1.5                                                         & 11.8 / 11.3                                                    & - / -                                & 33.2 / 92.2                          \\
                                                                           & NDT                                                            & 3.8 / 5.0                                                           & 1.2 / 2.5                                                      & 3.6 / 3.9                                                        & 5.0 / 12.3                                                       & 2.5 / 3.3                                                         & 2.1 / 2.3                                                         & 3.9 / 3.7                                                      & 3.4 / 4.5                            & 3.2 / 4.7                            \\ \midrule
            BA                                                             & BALM                                                           & 2.4 / 3.8                                                           & 0.6 / {\color{axesBlue}{\textit{0.7}}}                         & 1.1 / 1.7                                                        & 4.0 / 11.1                                                       & 1.7 / 2.5                                                         & 1.2 / 2.0                                                         & 2.8 / 3.2                                                      & 2.1 / 4.8                            & 2.0 / {\color{axesBlue}\textit{3.7}} \\
                                                                           & Ours                                                           & {\color{axesBlue}{\textit{1.1}}} / {\color{axesBlue}{\textit{1.8}}} & {\color{axesRed}\textbf{0.4}} / 1.0                            & 0.9 / {\color{axesBlue}{\textit{1.4}}}                           & {\color{axesBlue}{\textit{2.4}}} / {\color{axesRed}\textbf{5.8}} & {\color{axesRed}\textbf{0.8}} / {\color{axesRed}\textbf{1.0}}     & {\color{axesRed}{\textbf{0.7}}} / {\color{axesRed}{\textbf{0.8}}} &
            {\color{axesBlue}\textit{2.0}} / {\color{axesRed}\textbf{1.9}} & {\color{axesBlue}\textit{1.2}} / {\color{axesRed}\textbf{1.9}} & \color{axesRed}\textbf{1.2} / \color{axesRed}{\textbf{2.0}}                                                                                                                                                                                                                                                                                                                                                                                                                                                                                                       \\
            \bottomrule
        \end{tabular}
        \label{tab:eval_acc}
    \end{center}
\end{table*}

\subsubsection{Simulation setup}

We follow \cite{ferrer2019eigen} to design a simulator that randomly generated poses and planar features.
For each feature, the local observations in individual frames are also produced by the simulator, as a consequence of which the associations are pre-defined.
Besides, to show the advantages and disadvantages of different methods, we validate the performance of different methods with different parameter configurations, including the number of planes (\#planes), number of poses (\#poses), and noise of the observations ($\sigma$);
Given a group of observations generated with pre-set parameters,

For the comparison, we use an EVM-based approach Eigen-Factor (EF), and PL-based method LIPS as baseline methods.
Additionally, we re-formulate the constraint proposed by EF into a least-squares manner, which can then be solved using the second-order method (e.g., LM) for better convergence speed.
With Cholesky decomposition, we have $\mathbf{S}_i = \mathbf{L}_i\mathbf{L}_i^T$. Then \autoref{eq.ef} can be rewritten as a least-squares problem, given as:
\begin{equation*}
    \begin{aligned}
        E  
        = \sum_k \boldsymbol{\eta}^T \mathbf{T}_{k} \mathbf{L}_k \mathbf{L}_k^T \mathbf{T}_{k}^T \boldsymbol{\eta}
        = \sum_k \left\| \mathbf{L}_k^T \mathbf{T}_{k}^T \boldsymbol{\eta} \right\|_2^2.
    \end{aligned}
\end{equation*}
With this least-squares formulation, the Jacobian computation can be much efficient and the second-order methods (e.g., GN or LM) provides better convergence speed.
We denote this baseline as \textbf{EF(LM)}.





\subsubsection{Results and discussions}

The evaluation is illustrated in \autoref{fig.sim_err}. Generally, we observe that compared with other methods, the proposed formulation achieves the best performance considering both accuracy and speed.
For EF, we observe that, as stated in \cite{liu2020balm}, the first-order gradient method used by EigenFactor is inefficient to converge.
On the contrary, the second-order method is more efficient to converge.

Also, as LIPS does not consider the feature parameter estimation noises, we observe that the corrupted coefficient leads to inaccurate estimation results.
On the contrary, EF, EF(LM) and ours estimate the model coefficients after each iteration, which guarantees the coefficients to be optimal under current pose estimation.
In other words, such estimation results minimize the objective function with the current pose estimation results.
LIPS, however, performs parameterization on local observations, which is considered to lose certain information in the raw measurements.
This is efficient in some test cases.
However, as the model coefficients are estimated from local observations, it is not guaranteed to be optimal.

Besides, while LIPS is efficient when the number of features is small, with the number of features growing, LIPS becomes less efficient.
Considering a typical range-based registration problem, there are generally thousands of features, which would introduce a significant number of parameters to optimize.

\begin{figure*}[t!]
    \centering
    \includegraphics[width=0.95\textwidth]{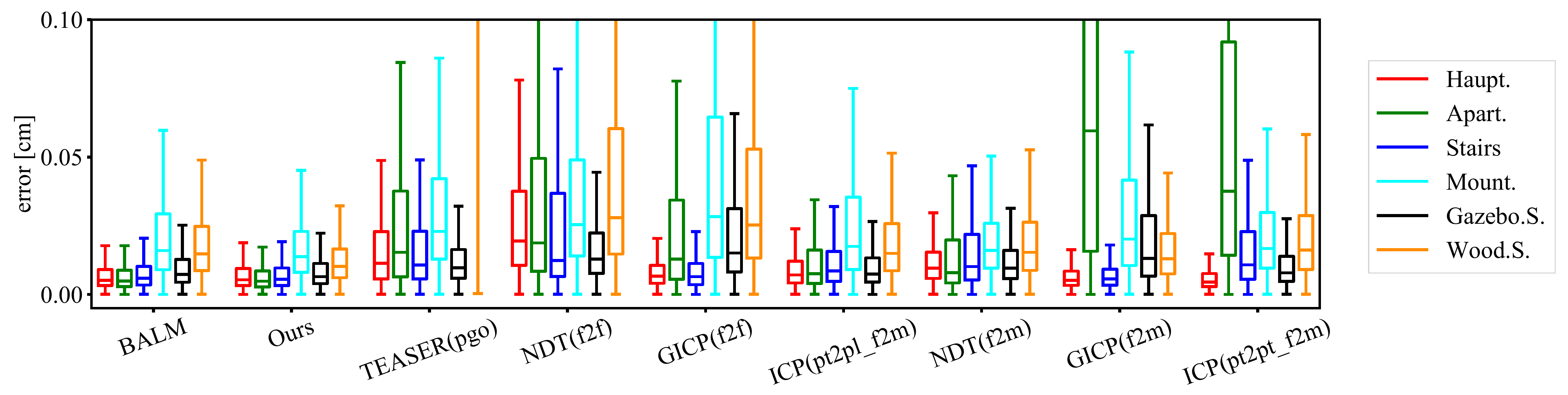} \,
    \caption{
        Evaluation on the structural accuracy of the reconstructed point cloud against the groundtruth.
    }
    \label{fig.str_acc}
    \vspace{-1em}
\end{figure*}

\begin{figure}[t!]
    \centering
    \subfloat[][ICP\_pt2pl(F2M)]{\includegraphics[width=0.23\textwidth]{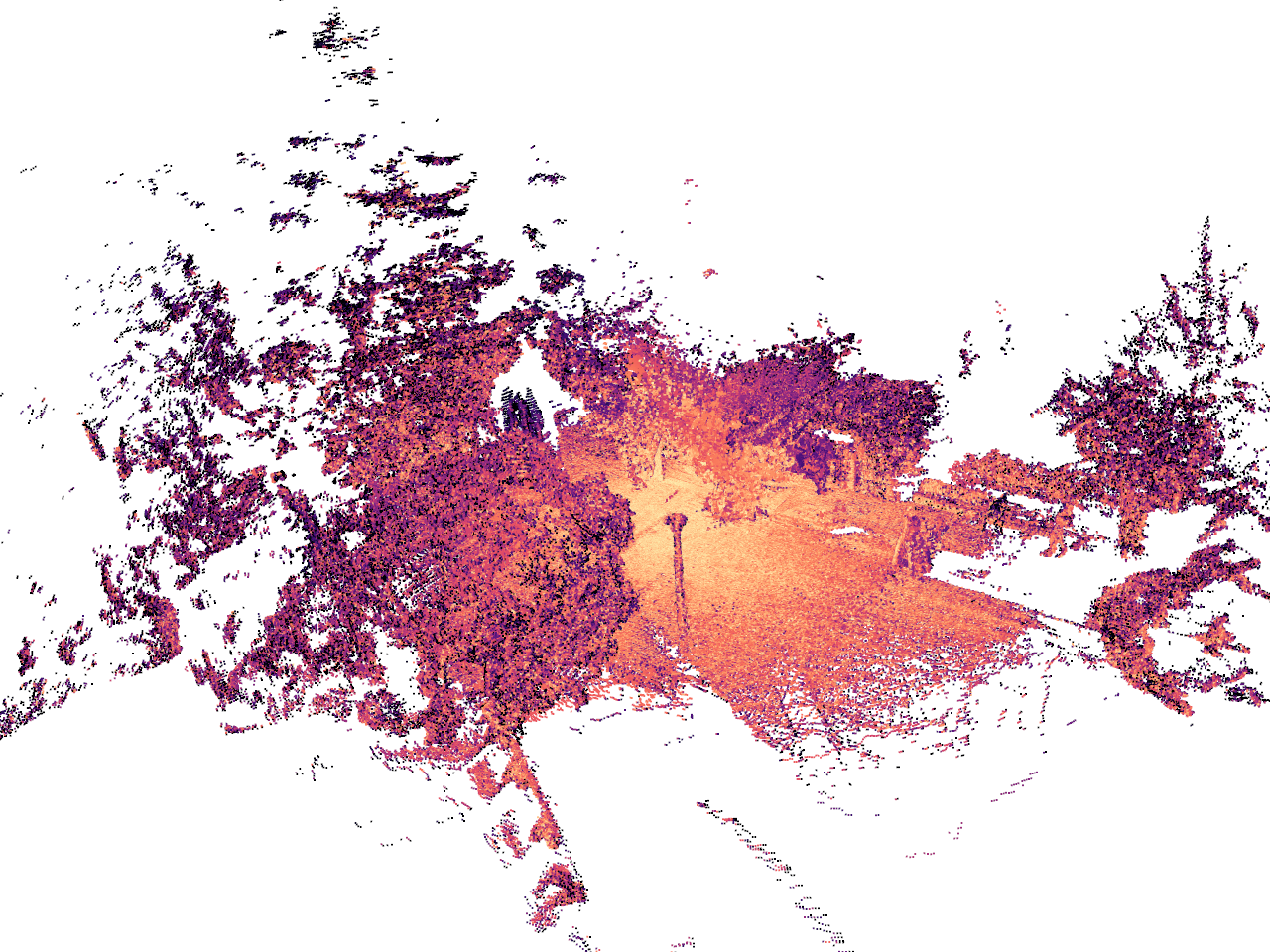}} \enspace
    \subfloat[][BALM]{\includegraphics[width=0.23\textwidth]{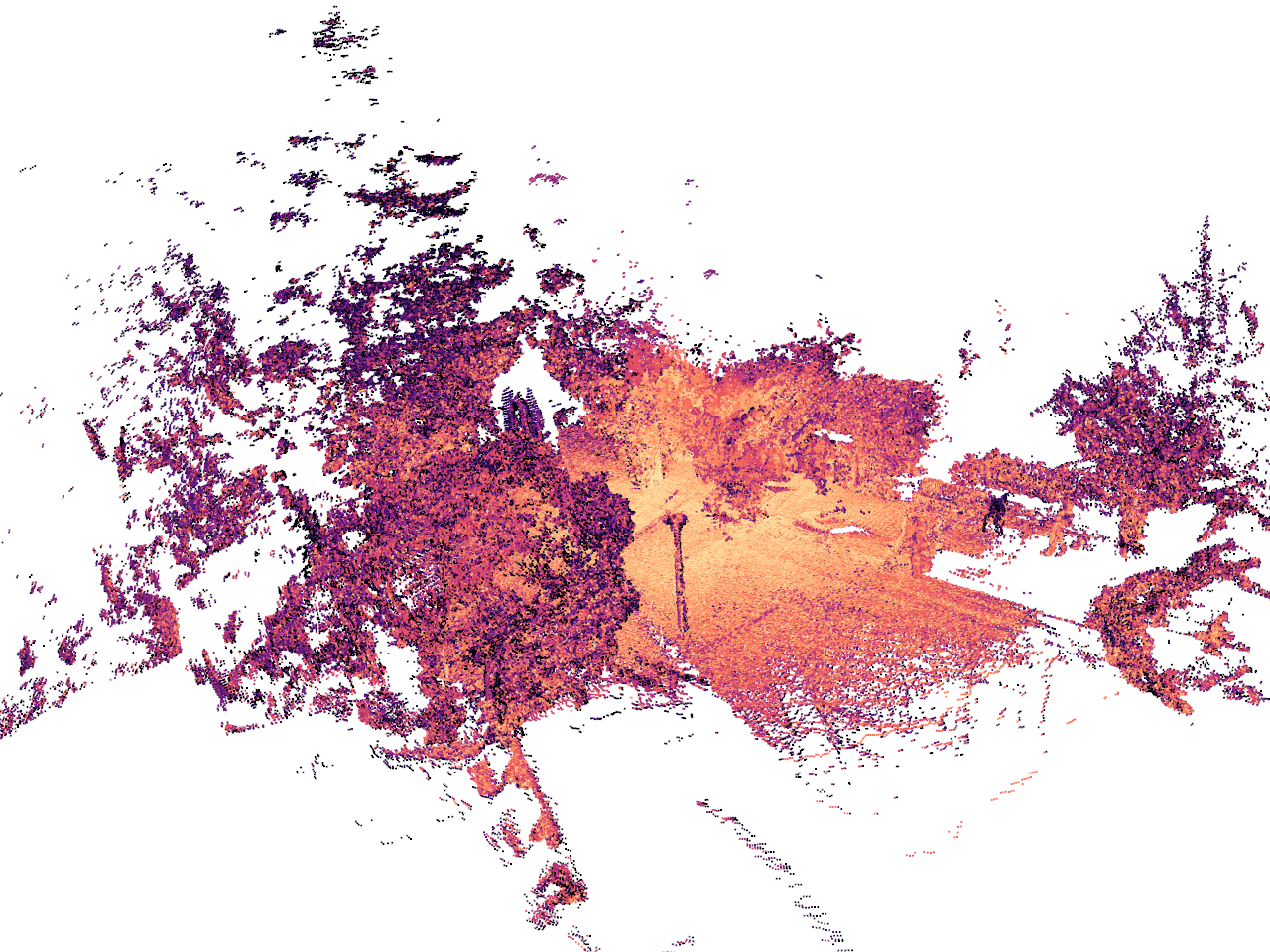}} \\
    \subfloat[][NDT(F2F)]{\includegraphics[width=0.23\textwidth]{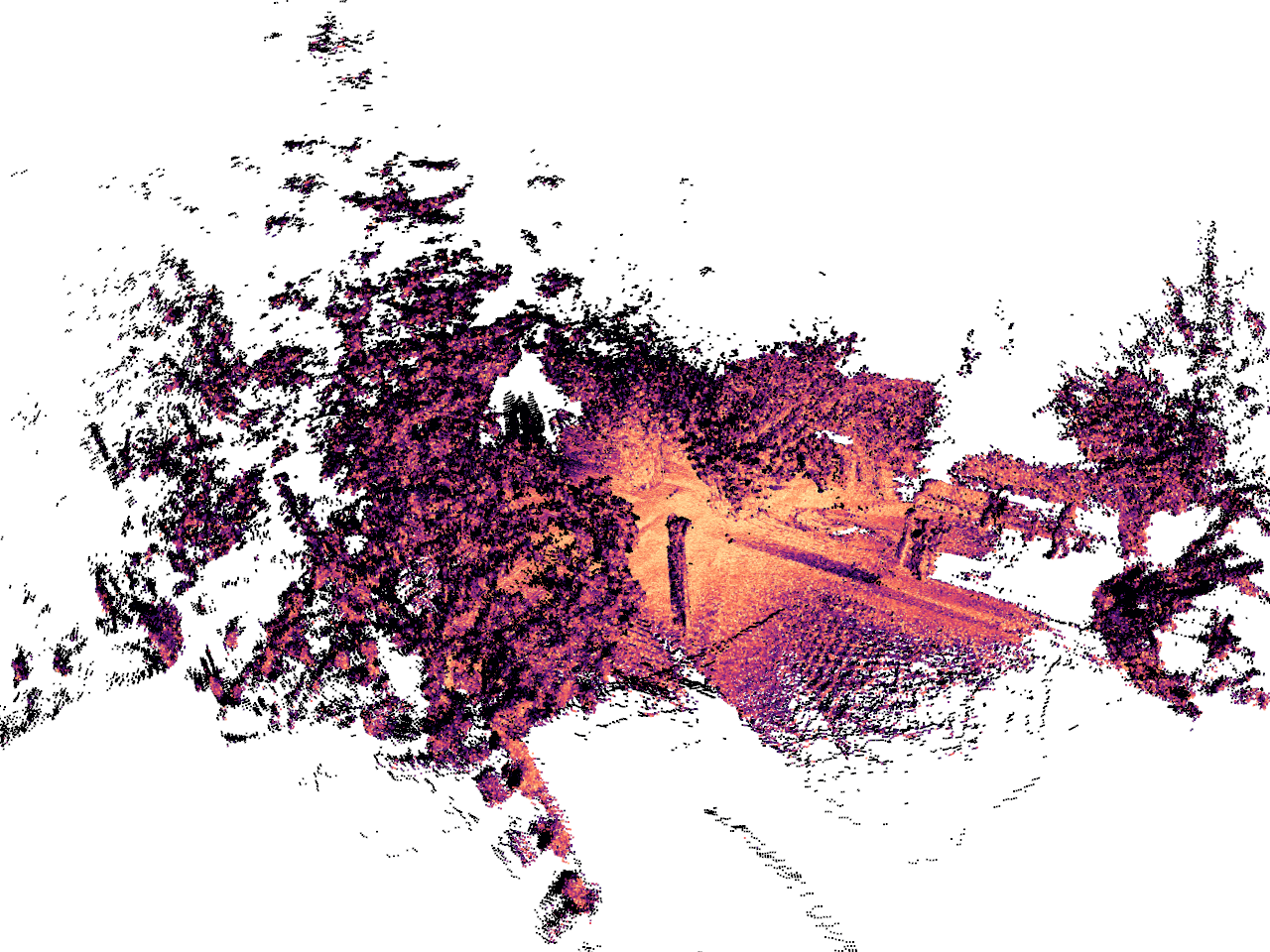}}\enspace
    \subfloat[][Ours]{\includegraphics[width=0.23\textwidth]{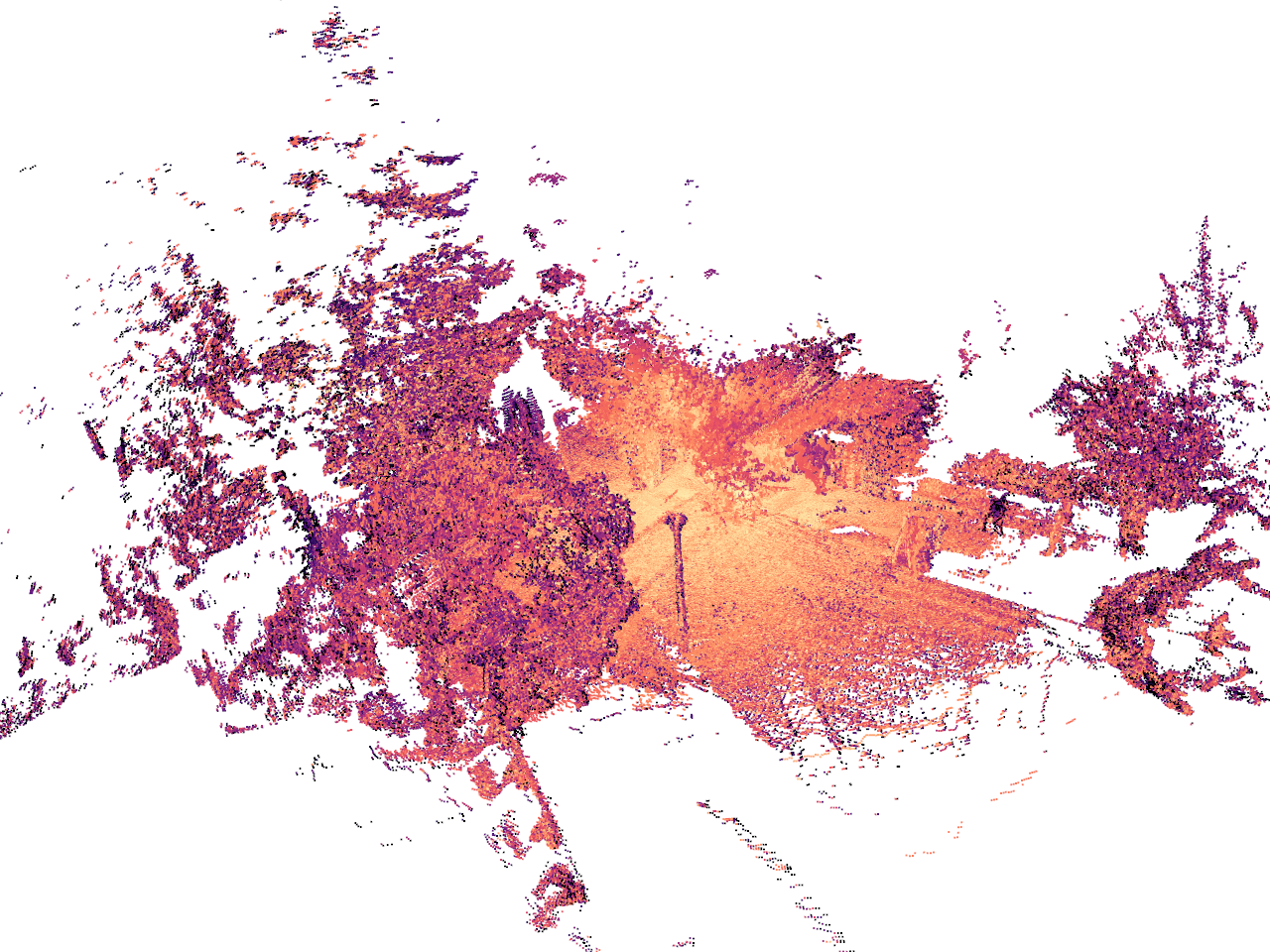}} \\
    \vspace{1ex}
    \includegraphics[width=0.48\textwidth]{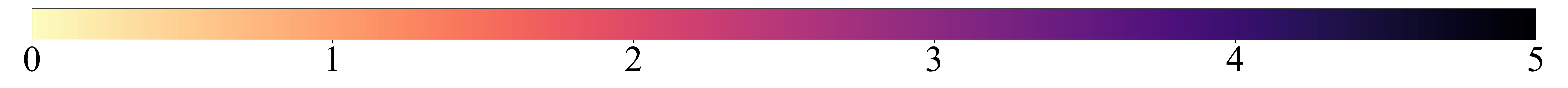}
    \caption{Heat maps showing the reconstruction error of 4 methods from a fixed view on sequence Gazebo.S. The range of error is set to 0-5 cm as showing in the bottom colorbar.
    }
    \label{fig.qualitative}
    \vspace{-0.5em}
\end{figure}

\subsection{Real World Experiemnts}

We perform experiments on the ETHZ Registration Dataset \cite{pomerleau2012}, a widely used dataset for registration covering both structured and unstructured scenarios.
Each sequence contains 30$\sim$45 frames of point cloud scans.
For the details of the dataset, we refer the reader to \cite{pomerleau2012}.
To validate our method, we evaluate methods in several categories for comparison, which are described as follows:
\textbf{Global} registration methods TEASER \cite{yang2020teaser} and FGR \cite{zhou2016fast}, which register consequential frames without initial guess;
\textbf{Global+Sync} methods, which extend global methods to utilize more inter-frame information. In the implementation, pairs of point cloud that share an overlap ratio over 60\% are registered by a global method.
Then, each method fine-tunes the initial pairwise estimations using standard Pose Graph Optimization (PGO);
\textbf{Frame-To-Frame} (F2F) methods that contain several ICP variants, which are most widely used for point cloud registration, including point-to-point ICP (ICP(pt2pt)), point-to-plane ICP (ICP(pt2pl)), NDT, GICP and VGICP;
\textbf{Frame-To-Map} (F2M) methods extending basic registration methods in a frame-to-map fashion, which better utilize inter-frame information. In the implementation, each scan is registered with first the previous scan and then the global map. After registration, the transformed scan is integrated to the global map;
\textbf{BA} methods including BALM and our method (Ours), which simultaneously optimize poses of several frames.






For F2F, F2M and BA methods, the estimations from TEASER are used as the initial guess.
We found that F2F and F2M methods occasionally fail when using raw measurements directly.
Therefore, for these methods, the input point cloud is downsampled with a resolution of 0.1 m.

\subsubsection{Evaluation of Registration Accuracy}

We evaluate the accuracy of the registered frame poses.
The evaluation metrics are the Relative Pose Error(RPE) and the Absolute Pose Error(APE) \cite{sturm2012benchmark}.
RPEs are computed by the estimated poses of two adjacent frames, while the APEs are calculated between the estimated trajectory and the groundtruth after translational alignment.


\autoref{tab:eval_acc} illustrates the evaluation of the registration accuracy.
Generally, by exploiting inter-frame constraints and jointly optimize the pose parameters, the average registration error of BA methods is on par with or better than other category of methods.
Our method achieves the best RPE (1.2cm) and APE (2.0cm) while BALM achieves the second best APE (3.7cm).
The RPEs of different methods are actually very close and there is only trivial or no improvement using BA.
For example, on sequence Wood.A., the RPE of ICP(pt2pl) and ours are 1.1cm and 1.2cm, respectively.
However, for the APE which represent the global consistency, our method is generally better than other baselines.
NDT(F2M) and VGICP(F2M) is very similar pipeline compared to ours in implementation.
From the results, we observe that while NDT(F2M) and VGICP(F2M) have accumulated drifts and inevitable estimation inaccuracy (even fails in some cases), our method well maintains the global consistency in all the sequences.
This verifies the benefit of using BA in multiview registration.




\begin{figure}[t!]
    \centering
    \includegraphics[width=0.48\textwidth]{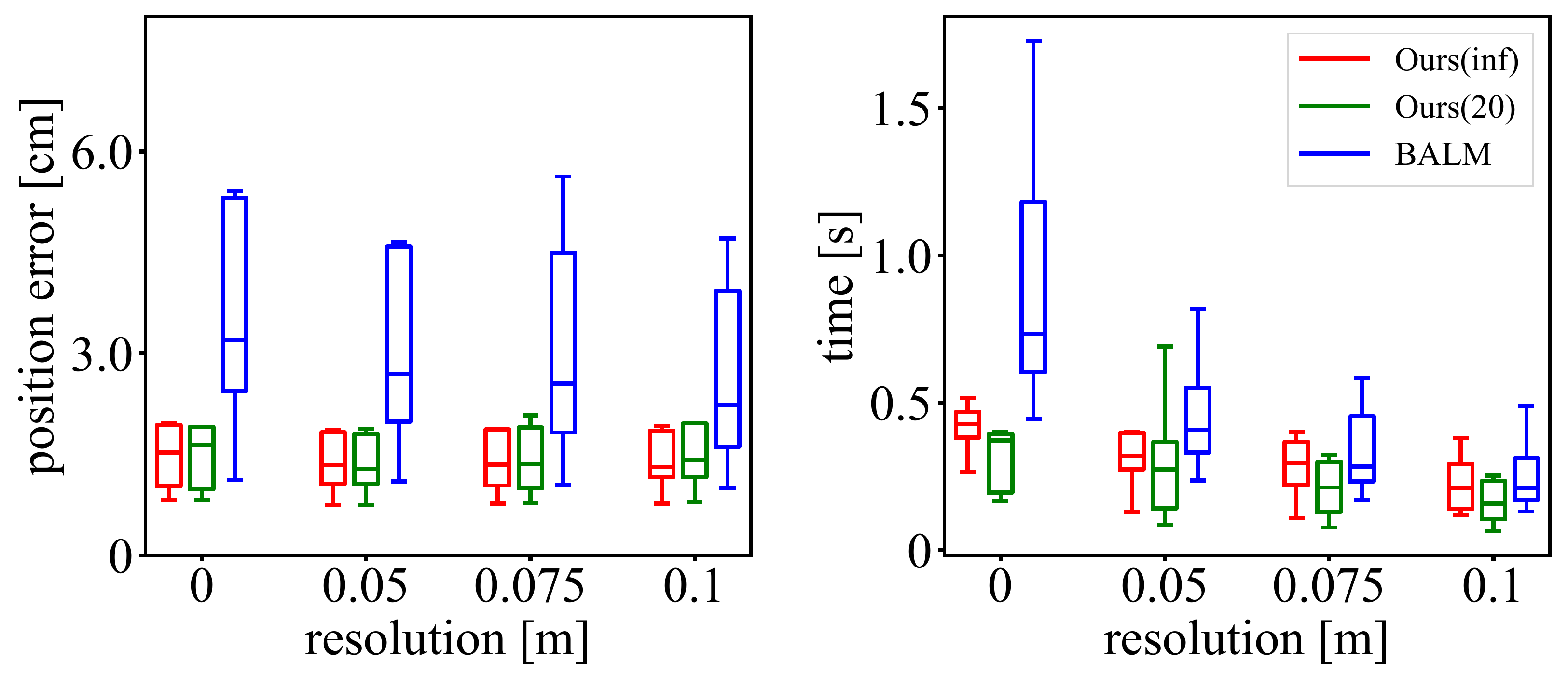} \\
    \caption{Evaluation on the translational error and runtime w.r.t. the resolution of the input point cloud.}
    \label{fig.comp_balm}
\end{figure}

\subsubsection{Evaluation of the Reconstruction Quality}

We concatenate the point cloud with the estimated poses for a global map, and then align it with the 3D model from the groundtruth.
The structural error is evaluated by calculating the distance between each point and its closest neighbor from the groundtruth geometry.
\autoref{fig.str_acc} shows the distribution of structural error on 6 representative sequences.
The results show that our reconstruction is on par with or better than the existing method.
The structural error is generally caused by the drift in pose estimation,
therefore these qualitative results also confirm the quantitative evaluation in \autoref{tab:eval_acc}.
\autoref{fig.qualitative} visualizes the reconstruction quality of our method and 3 other representative methods on Gazebo.S. sequence.
Despite the unstructured environment, our method works well and the reconstructed point cloud well aligns with the ground truth.
In addition, we observe that with our formulation, the regions with large structural error are generally non-planar cases.
This indicates that more generic model would contribute to the registration performance.

\subsubsection{Comparison with BALM}
We further perform detailed comparison on registration accuracy and runtime with BALM, and the results are shown in \autoref{fig.comp_balm}.
We adopt two variants of our method, denoted as Ours(20) and Ours(inf), where the number of optimization iterations are set to 20 and infinite (iterating until convergence), respectively.
With the variances of input resolution, the position error of each method is close and the runtime of our methods does not change significantly.
In contrast, the runtime of BALM increase a lot when directly using the raw measurements.
This is consistent with our observation in \autoref{sec.bg} that the complexity of BALM is dependent of the number of input point cloud, while that of our method is dependent of the number of features.
Originally, we expect the runtime of our method is constant when the resolution changes.
However, in the experiment, we found that the changes affect the association to some extend,
as a consequence of which the runtime of our method also increases if the resolution of input data is high.




\section{Conclusions}
\label{sec.conclusion}

In this work, we have reviewed prior arts on the problem of multiview registration in detail, especially for PL-based and EVM-based methods.
To introduce our formulation on this problem, we have first provided a theoretical analysis on the EVM-based formulation's optimal condition, yielding that it can be uniformed with PL-based methods in a noise-less situation.
Then, we have introduced a different objective function that weighs rotational, and translational terms by the eigenvalues from decomposition to handle the measurement noise and the computational cost properly.
Finally, we have proposed a multiview registration system that utilizes the above formulation, voxel-based data management for feature association and local distribution aggregation for optimal state calculation.
Both simulation and the real-world experimental results validate the proposed method.

Our current implementation focuses on multiview point cloud registration, which leaves SLAM problem with sparse LiDAR scans unexplored.
We consider that it would be meaningful to apply the proposed method in a range-based SLAM system and perform further analysis.
In the future, we would like to investigate more effective approaches in data quantization further.
Unlike some KD-Tree-based methods, the association step relies on the initial estimation in the current voxel-based implementation.
Although it is highly efficient, it is supposed to be more sensitive to the local minimum.
This is significant to the feature association for state estimation.





%

\bibliography{main}
\bibliographystyle{IEEEtran}

\addtolength{\textheight}{-12cm}   

\newpage
\includepdf[page=1-]{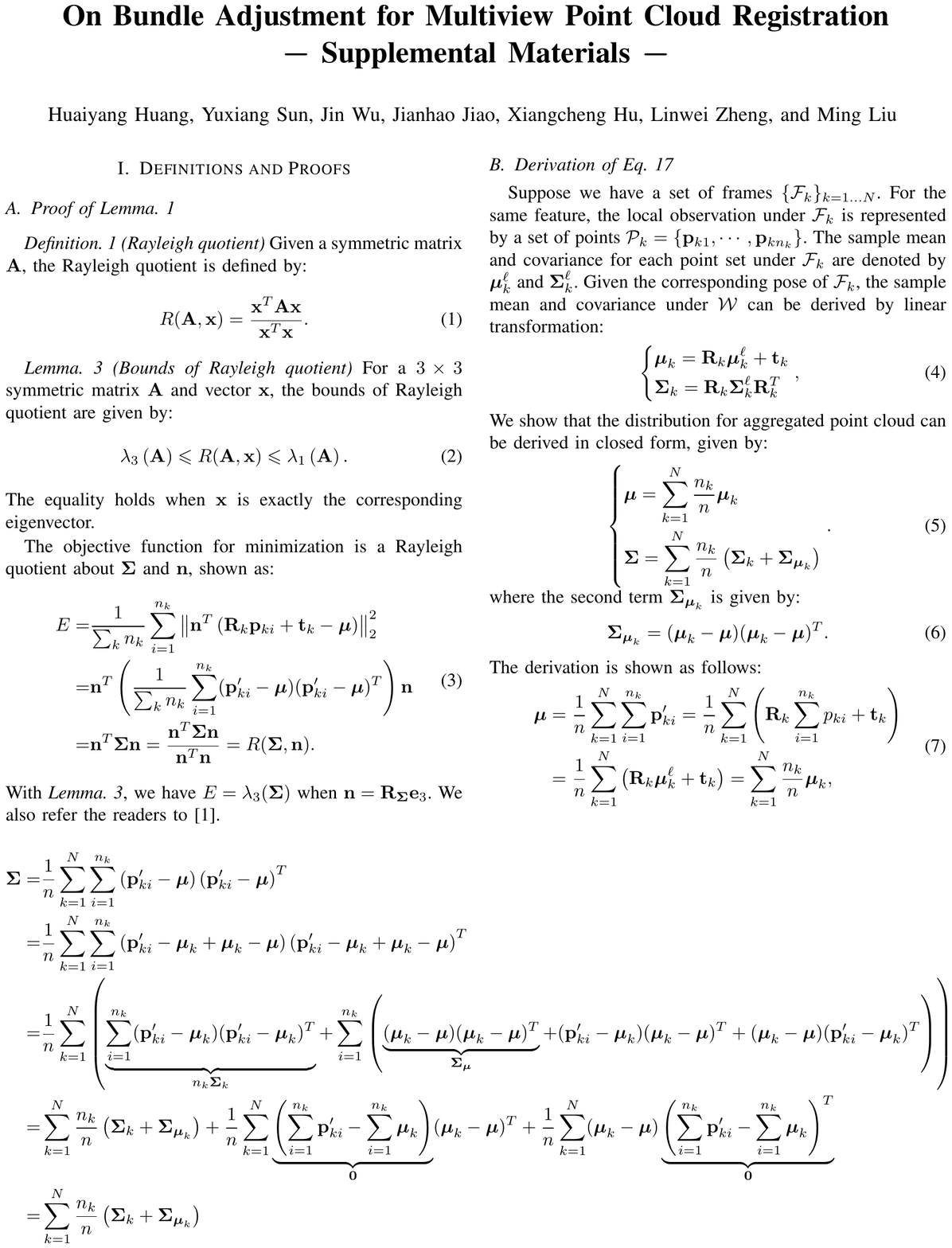}

\end{document}